# LERIL : Collaborative Effort for Creating Lexical Resources

Akshar Bharati,
Dipti M Sharma, Vineet Chaitanya, Amba P Kulkarni, Rajeev Sangal
Language Technologies Research Centre
International Institute of Information Technology Hyderabad
*{dipti,vc,amba,sangal}@iiit.net*

Durgesh D Rao
National Centre for Software Technology, Mumbai
*durgesh@ncst.ernet.in*

## *Abstract*

The paper reports on efforts taken to create lexical resources pertaining to Indian languages, using the collaborative model. The lexical resources being developed are: (1) Transfer lexicon and grammar from English to several Indian languages. (2) Dependencey tree bank of annotated corpora for several Indian languages. The dependency trees are based on the Paninian model. (3) Bilingual dictionary of 'core meanings'.

## 1. Introduction

Non-availability of lexical resources in the electronic form is a major bottleneck for anyone working in the field of NLP on Indian languages. It was decided to take some measures which would remove this bottleneck in a quick and efficient way.

As a first step in this direction a collaborative effort was undertaken to develop a bilingual electronic dictionary in the open source model. The interesting aspect of this effort was that the work was carried out by school children, teachers, housewives, and retired people among others. People in about 8 cities were involved in the exercise. The school teachers participated, to some extent, in correcting and refining the work. This was later edited by a small core team of two.

The development of the dictionary resource took advantage of the bilingual ability of the contributors. The contributors provided the basic data:

a) A number of Hindi equivalents required to cover various senses of the English lexical item in different contexts.

b) An English example sentence for every Hindi equivalent.

The developed resource is now available as an "open resource" under General Public License. ( GPL,1991 )

It might appear difficult to create a major resource like a dictionary in this way, with a diverse set of people working on it. Admittedly there are variations in quality at present. But the coverage is already quite exhaustive. A number of factors, however, made it possible:

1. The contributors were advised to consult various mono- and bi-lingual dictionaries. Many contributors, including students working in a classroom setting with a local teacher, consulted monolingual advanced learner's dictionaries for English. (However, they did not copy the entries (which anyway were in English alone), instead they supplied Hindi equivalents for the available detailed differentiation of English senses, wherever the Hindi equivalents were different, or represented different meanings in their judgement.)

2. The initial information that was to be incorporated in the dictionary was kept to a minimum so that anyone who is sufficiently bilingual could participate in the activity. This is why even the school children could contribute to the effort.

3. Some amount of editing was carried out by a small central team. (However, in future we would like this also to be carried out in a distributed way, perhaps out of a few tens or hundreds of better trained people selected out of hundreds who participated in the initial exercise. Only the final output would be corrected by the small centralized team.)

Modern technology permits the incremental improvement and enhancement of the basic resource over a period of time. This was a basic consideration in embarking on such an exercise. The result of this effort has led to the rapid creation of the present dictionary (Shabdaanjali; 2000). Which is available as an open resource under General Public License (GPL,1991.) The dictionary consists of more than 25000 headwords, with fairly detailed differentiation of senses.

Here is an example entry from Shabdaanjali which gives the senses as well as example sentences illustrating the senses:

"go","V",
--"1.jAnA"
I go to school.
--"2.rakhA~jAnA"
These clothes go into that suitcase.
--"3.samAnA[<jAnA]"
This key will not go in that lock.
--"4.calanA"
How did the meeting go?

--"5.ho~jAnA{sthiti}"
Have you gone mad?
--"6.aAvAjZa~honA/karanA"
The bell has gone for this period.
--"7.nikala~jAnA"
The P.M. has already gone.

Notice the level of details in sense differentiation.

Once the task (Task 0) of building a basic dictionary was over, it was decided to go ahead and build some more specific resources based on recommendation of LRNLP 2001.

**2. Tasks on hand**

The next level of work, based on the above resource is being carried out by several sites. Some of the work is voluntary, and some are using locally available financial resources to pay for the work. Three tasks are being carried out at present are:

The tasks chosen are:
1. TransLexGram (Transfer Lexicon and Grammar),
2. AnnCorra: Annotated corpus for each Indian language,
3. Shabda-Sutra - Bilingual dictionary of core meaning with the above applications in mind.

**2.1 Principle for Selection of Tasks**

The principle used in selecting these tasks is that: LRs must be directly useful for at least one target application.

- The application provides focus and prioritizes tasks.

- Restricts scope of difficult problems. Instead of general solution, special solutions or work- arounds are alright.

- Allows common decisions among teams. Acts as a yardstick to decide among alternatives.

## 2.2 Target Applications

The target applications which were selected were:

- Machine translation or language access
  - Among Indian languages
  - From English to Indian languages

- Multilingual information retrieval

## 3. Task 1 : TransLexGram (Transfer Lexicon and Grammar)

The first task being attempted, based on the Shabdaanjali is to produce transfer lexicon and grammar from English to Hindi.

The task at this level has two components:
 - Providing translation of the English sentences into Hindi.
 - Creating a parallel grammar from English to Hindi.

For example, the developer is provided with the entry for word 'go' from Shabdaanjali in a machine readable format (only part of the entry is given below for illustration). He then fills in the above two components of information. At the end, the entry has the following additional information -

HEADWORD::"go","V"

MEANING::1::"jAnA"
ENG_EXP:: I go to school.
TR_NAT:: maiM skUla jAtA hUM.
TR_ENG-INFLNC::
FRAME_E:: A goes to B
FRAME_I:: A B [ko] jAtA hai
ERR::
COMNT::

MEANING::2::"rakha~jAnA"
ENG_EXP:: These clothes go into that suitcase.
TR_NAT:: ye kapaDe usa sUtakesa meM rakhe
         jAyeMge
TR_ENG-INFLNC::
FRAME_E:: A goes into B
FRAME_I:: A B meM rakhA_jAtA_hai
ERR::
COMNT::

The contributors are requested to follow the following scheme while translating -

a) To give natural translations in their respective language. This has to be entered in the the field TR_NAT.

b) In case an alternative translation having English influence is possible, it has also to be entered in the field TR_ENG_INFLNCE.

c) The option of giving more than one translation of the same sentence is also given. These are to be entered by repeating TR_NAT field.

d) For contributors for languages other than Hindi, the field MEANING_OTH is provided where they provide the sense in their respective language.

The second componenet calls for the verb frames to be in a standard format for all verbs, so that these can be semi-automatically processed. The contributors are, therefore, provided with a guideline which clearly states that irrespective of the mood and tense of the example sentence, the frames should be in simple present tense (habitual present tense) only. (An expansion of the field names is attached as Appendix-1)

The development of Transfer Lexicon and Grammar is being done from English to several Indian languages. All of them are using the dictionary Shabdaanjali as a starting point.

There are six languages for which this work is being carried out. The languages apart from Hindi are, Tamil and Telugu from southern India, Marathi and Gujarati from the western region and Oriya from the east.

As indicated above, we try to define verb frames for English and the corresponding post-position for each Indian language. Linguists are the preferred people for doing this task, but the transfer grammar is kept so natural that a person without linguistic training but with language sensitivity can also contribute to it after a short period of training or self-learning is needed. Though this grammar is expected to be used by the machine directly, the framework has been kept rather simple.

The effort for Transfer Lexicon and Grammar will serve several purposes;

(a) It will generate bilingual dictionaries with detailed differentiation of senses, one for each of the Indian languages above. Such online bilingual dictionaries are usable by human beings. They can also be the stepping stones for machine translation dictionaries. (In fact, Shabdaanjali: English-Hindi dictionary in its present form without any change is being used by the experimental anusaaraka machine translation system.)

(b)  The example sentences are an important aid in future development and refinement of the dictionary.  Our experience shows that without the example sentences, lexicographers find it very hard to understand and refine senses in a given dictionary.

(c) It will generate corpora of parallel sentences in English and several Indian languages. Each parallel corpus will consist of sentences that are examples which differentiate among the various senses of a word.  Thus, it is a special contrastive set, which will be of special value and might also be used for developing systems across Indian languages.  Such a parallel corpus along with machine learning techniques can be used to produce example based machine translation system.

(d)  The transfer patterns given in Frame_E and Frame_I have been kept simple so that even ordinary bilingual speakers can provide the information.  However, these can be directly used by a machine translation system.  (In fact, a number of experiments have already started using the above.)

The interesting part is that the work is being done by a loosely knit set of sites, institutes, and individuals. Technical coordination is done by a set of senior researchers.

## 4. Task 2: AnnCorra: Annotated corpus for each Indian language

Second task pertains to building a "tree-bank" or tagged corpus, in which the tags indicate the sentential analysis. The purpose behind this effort is to fill the lacuna in such resources for Indian languages. It will be an important resource for the developement of Indian language parsers, machine learning of grammars, lakshan charts (discrimination nets for sense disambiguation) and a host of other tools.

Dependency analysis using the Paninian theory, is being used to mark the sentential analysis for sentences in each language (Bharati et al, 1995), since the Paninian grammatical model has been chosen for sentence analysis. Preference for this model is based on the following factors -

1) The model was developed  based on Sanskrit.  Hence, it can deal better with the constructions in Indian languages.

2) The model offers unique methods for analyzing syntactico-semantic  information, and would also be eminently suitable for analyzing other languages of the world.  (The analysis offered is a kind of dependency analysis with principled answers to the problems of case- based grammars.)

*For example -*

- rAma ne phala kATakara pAnI piyA.
- Ram -ne fruit having-cut water drank
- (Having cut fruit, Ram drank water.)

Analysis tree for the above sentence is:

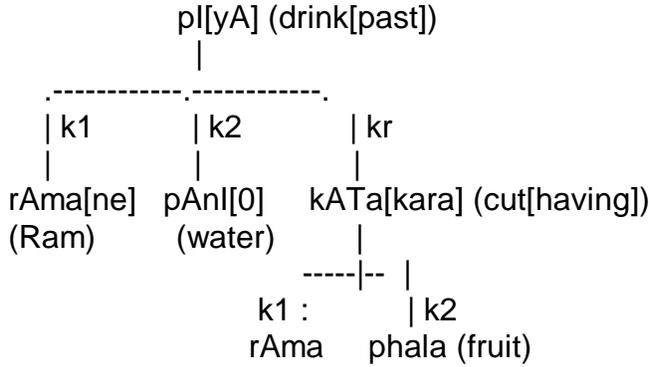

Where k1 is karta karaka, k2 is karma karaka from the Paninian theory (Bharati et. al, 1995, Chapter 5).

Linear notation for the above tree is:

- rAma_ne/k1->i  phala/k2->j  kATakara/kr:j->i
- Ram_erg.        fruit          having-cut

    pAnI/k2->i   piyA::v:i
    water        drank

This linear notation is powerful enough to represent arbitrary dependency trees. Several defaults are used to reduce typing as shown below.

- rAma_ne/k1->i  phala/k2  kATakara/kr
- Ram_erg.  fruit    having-cut

    pAnI/k2  piyA::v:i
    water    drank

For example, the word 'phala' marked by k2 karaka by default attaches to the nearest verb 'kATakara'.

This work requires a higher level of expertise, and is planned to be done in a collaborative model by Sanskritists and other linguists trained in Paninian analysis. It is expected to draw people trained in the traditional Sanskrit shastras, particularly vyakarana shastra. (grammar).

For maintaining the consistency, the same TAGSETS are provided to all the contributors. Also, most of the annotators come from a strong tradition of Sanskrit grammar. Which provides principled analysis of syntactico-semantic issues, thus resulting in consistent output.

## 5. Task 3: Shabda-Sutra - Bilingual dictionary of core meaning

The third task involves semantic analysis of words across languages. Polysemy is a major problem that one has to deal with while building bilingual lexical resources for the machine translation. The concept of 'Shabdasutra' is an attempt to capture the underlying thread which relates various meanings in a polysymous word.

The term 'sutra' in 'Shabdasutra' is used at two levels.

**A.** At the first level the term 'Shabdasutra' means 'a formula' which encodes the basic semantic concept of a word and how it gets extended to varying usages. For example. The English word 'issue' has several meanings as available from Shabdanjali: Shabdasutra or formula for the English word 'issue' is

viSaya[~~ < niSpAdana]

or its rough gloss:

topic[~~ < to come into existence]

Notational symbol '<' means 'is derived from' and symbol '~' means that the sense has taken several turns in its evolution. Thus, the above notation says that the meaning of 'issue' is 'topic' which has arisen from 'to-come-into-existence'after taking many turns in its evolution. This 'sutra' is a formula which expresses that 'niSpAdana' appears to be the basic sense or the 'core' meaning of the English word 'issue'. From this 'core sense' various other meanings have evolved.

**B.** The second sense in which the term is used is that of an 'underlying thread' which connects all the senses to which the meaning of a particular word gets extended. To continue the example of 'issue' above, the formula given above has the following underlying thread :-

niSpAdana(astitwa meM lAnA/AnA)
 --> niSpatti kA srota
--> niSpatti (santAna, sansakaraNa etc)
which means:

'bring into existence
 ---> point of origin
---> the thing that comes into existence (child,
      edition etc).

The relation between various senses of the word 'issue' can be seen through this 'sutra' with the help of following examples from english

niSpAdana  eg:  "issue orders"

-->niSpatti kA  srota   eg:  "point of issue of a
                                           river"
-->niSpatti  eg : "has no issue after marriage,
                 latest issue is out,"

The way the 'underlying thread' is compressed into a 'sutra(formula)' notationally can vary depending on the complexity of the sense it is encoding.

Following are the steps in this task:

- Begin with a bilingual dictionary of English to Indian Languages which contains different senses, and example sentences for each sense
- Identify commonality of meanings for a word
- Come up with core meaning or word-thread or  sabda-sutra

This is an intricate task, and has been completed by a group of dedicated researchers for 5000 words.

For all the above tasks, a basic list of 5000 words based on high frequency is being used. The initial target is to complete 5000 high frequency words for all the Indian languages. In case, some group  wants to go further and work for a larger dictionary they can cover the whole dictionary (with about 25000 Headwords).

??The target is to complete the first phase work (5000 words) in several Indian languages by the end of November, 2001.??

**6. Policy for Distribution**

The resources so developed would be available to people at no cost or low cost. These are like infrastructure, which everyone uses, but finds  difficult to pay for.

Most importantly, the above resources would be "open source" under GPL. This is to allow others to work on the resource, modify or refine it, and then redistribute it.

## 7. Conclusions

This paper reports on some efforts which have created or are creating lexical resources pertaining to Indian languages, using the voluntary collaborative model. One of the novel idea was to involve several hundred school children spread over several cities, to yield a detailed bilingual dictionary, which is now not only available for consultation by the general public, but is also being used as a stepping stone for building several other kinds of lexical resources namely, (1) Transfer lexicon and Grammar, (2) Annotation of Corpora and (3) Bilingual Dictionary of core senses These resources are being developed with machine translation and information retrieval in mind. The lexical resources so produced will be distributed as "open" or "free" resources under GPL.

## Acknowledgements

The frameworks for TransLexGram and AnnCorra in the result of discussions with several people such as: Prof. Aravind K Joshi,
Dr. B. Srinivas, Dr. K.V. Rama Krishnamacharyulu Dr. Thakur Dass, Dr. V.P. Jain , among others. Many from LRNLP-2001 contributed to the framework through discussions. From that Workshop resulted the LERIL effort: Lexical Resources for Indian Languages.

**Appendix -I: 'FIELD NAMES' Provided in Task 1 (TransLexGram):**

HEADWORD - The lexical item for which the entry is being made
MEANING - Indian language equivalent for the Headword.
ENG_EXP - Example sentence in English
TR_NAT - Natural translation
TR_ENG-INFLNC - Translation having english influence
FRAME_E - Frame for the English sentence
FRAME_I - Frame for the Indian language translation

ERR - Error (this column is for human use)
COMNT - Comment (this column is for human use)

**Appendix -II: Tagsets**

The tagsets used here have been divided into two categories -

1) TAGSET-1 - Tags which express relationships are marked by a preceding '/' .
   For example karakas are grammatical relationships, thus they are marked '/k1', '/k2', '/k3' etc.

2) TAGSET-2 - Tags expressing type of node are marked by a preceding '::' Verbs etc. are nodes, so they will be marked '::v',

*Some example tags -*

TAGSET-1 (Expressing relationship labels) Marked '/'

 s     : Sentence
        Example -
        [rAma ne khIra khAyI]<s>
        [rAma postp milk-rice ate_fem]

 k1    : karta
        Example -
        [rAma_ne/k1 khIra khAyI]<s>

 k2    : karma
        Example -
        [rAma_ne khIra/k2 khAyI]<s>

k3    : karana
        Example -
        [rAma_ne cammaca_se/k3
        [rAma_postp spoon_with/k3

         khIra khAyI]<s>
         milk-rice ate_f]

TAGSET-2 (for nodes) Marked '::'

 v     : Verb

 Kr    : Gerund

 vH    : Verb-BE
         Example -
          rAma adhyApaka HE
           Ram   teacher   is

 yo    :Conjunct

The total number of tags is around 35. Since the task is going on, this may be revised, in case it is needed.